# Entailment in Probability of Thresholded Generalizations

Donald Bamber*
Naval Command, Control & Ocean Surveillance Center
Research, Development, Test & Evaluation Division[†]

## Abstract

A nonmonotonic logic of thresholded generalizations is presented. Given propositions $\alpha$ and $\beta$ from a language $\mathcal{L}$ and a positive integer $k$, the thresholded generalization $\alpha \Rightarrow^k \beta$ means that the conditional probability $\pi(\beta|\alpha)$ is at least $1 - \psi\delta^k$. A two-level probability structure is defined. At the lower level, a *model* is defined to be a probability function on $\mathcal{L}$. At the upper level, there is a probability distribution over models. A definition is given of what it means for a collection of thresholded generalizations to entail another thresholded generalization. This nonmonotonic entailment relation, called *entailment in probability*, has the feature that its conclusions are probabilistically trustworthy meaning that, given true premises, it is improbable that an entailed conclusion would be false. A procedure is presented for ascertaining whether any given collection of premises entails any given conclusion. It is shown that entailment in probability is closely related to Goldszmidt and Pearl's System-$Z^+$, thereby demonstrating that System-$Z^+$'s conclusions are probabilistically trustworthy.

## 1  INTRODUCTION

### 1.1  GENERALIZATIONS

By a *generalization* is meant any statement of the type: *Nearly everything that has property $\alpha$ also has property $\beta$*. Examples: (a) *Nearly every animal that has feathers is capable of flight.* (b) *On nearly every occasion when I turn the key in my car ignition, the car starts.*

In a logic of generalizations, collections of premises which are generalizations entail conclusions which are also generalizations. Why is a logic of generalizations needed? Why does one need to infer new, previously unknown, generalizations from already known generalizations? The reason is that the known generalizations may not be adequate to give unambiguous answers to questions about individual objects.

For example, suppose that, among the known generalizations, are:

(i) Nearly everything that has property $\alpha$ also has property $\gamma$.

(ii) Nearly everything that has property $\beta$ also has property not-$\gamma$.

If all that is known about an object $o$ is that it has property $\alpha$, then it should be concluded that, in all probability, $o$ has property $\gamma$. But what if it is known that $o$ has both property $\alpha$ and property $\beta$ (and nothing more is known about $o$)? What should be concluded then?[1] By the principle of total evidence (Schurz, 1994), the conclusion that $o$ has property $\gamma$ should be accepted as probably true if and only if it is known that:

(iii) Nearly everything having properties $\alpha$ and $\beta$ also has property $\gamma$.

Thus, it is desired to know whether (iii) is true, but (iii) is not one of the known generalizations. This illustrates why a logic of generalizations is needed. The generalization (iii) should be judged to be true if and only if it is entailed by the known generalizations.

### 1.2  THRESHOLDED GENERALIZATIONS

The *exception rate* for a generalization *Nearly everything that has property $\alpha$ also has property $\beta$* is defined to be one minus the conditional probability of $\beta$ given

---

*Address:* NCCOSC RDTE DIV 44215, 53355 RYNE ROAD RM 222, SAN DIEGO CA 92152-7252, USA. *Electronic Mail:* bamber@nosc.mil

[†]Usually abbreviated as *NRaD*.

[1]This is a version of the reference class problem (Bacchus, Grove, Halpern & Koller, 1992).

$\alpha$. A generalization is said to be *thresholded* if it is accompanied by the the information that its exception rate is less than some particular threshold.

It is useful to threshold generalizations for two reasons. First, it helps in gauging how much confidence to place in inferences about individual objects. Thus, suppose that all that is known about an object $o$ is that it has property $\alpha$. Given the generalization *Nearly everything that has property $\alpha$ also has property $\beta$*, how confident should one be that $o$ has property $\beta$? The answer is that, the smaller the generalization's threshold, the more confidence one may have. Second, the thresholds attached to premises affect what conclusions are entailed by them. For example, whether or not the generalizations (i) and (ii) entail the generalization (iii) depends upon the thresholds attached to (i) and (ii). If (i)'s threshold is much smaller than (ii)'s threshold, then (iii) will be entailed; otherwise, it will not be entailed.

A nonmonotonic logic of thresholded generalizations will be described in this paper.

## 2  SYNTAX AND SEMANTICS

In this paper, object properties will be represented by propositions in a propositional language.

Let $\mathcal{L}$ be a propositional language constructed from the primitive propositions $\sigma_1, \ldots, \sigma_r$ and the connectives *and* $\land$, *or* $\lor$, *not* $\neg$, and possibly additional connectives. Let $t$ be an abbreviation for $\sigma_1 \lor \neg\sigma_1$ and let $f$ be an abbreviation for $\sigma_1 \land \neg\sigma_1$. An *atom* is a proposition of the form $\varsigma_1 \land \cdots \land \varsigma_r$ where each $\varsigma_i$ is either $\sigma_i$ or $\neg\sigma_i$. Thus, there are $2^r$ atoms in $\mathcal{L}$. Let $\alpha \models \beta$ mean that $\alpha$ entails $\beta$ and let $\alpha \models\!\!\!\dashv \beta$ mean that $\alpha$ and $\beta$ are equivalent.

Because propositions in $\mathcal{L}$ represent properties of objects, they will be called *properties*. Thus, an object has property $\alpha \land \beta$ if and only if it has property $\alpha$ and property $\beta$. The properties $\alpha \lor \beta$ and $\neg\alpha$ are interpreted analogously. Furthermore, the property $t$ is the universal property possessed by every object and the property $f$ is the impossible property possessed by no object.

Suppose $\pi$ is a probability function on $\mathcal{L}$. Thus, if $\alpha \in \mathcal{L}$, $\pi(\alpha)$ may be interpreted as the probability of a randomly selected object having property $\alpha$. For any $\alpha, \beta \in \mathcal{L}$, define

$$\pi(\beta|\alpha) = \pi(\alpha \land \beta)/\pi(\alpha) \quad \text{if } \pi(\alpha) > 0 \quad (1)$$

and

$$\pi(\beta|\alpha) = 1 \quad \text{if } \pi(\alpha) = 0.\ ^2 \quad (2)$$

Also, define

$$\pi^*(\neg\beta|\alpha) = 1 - \pi(\beta|\alpha).$$

Then, $\pi^*(\neg\beta|\alpha)$ is the exception rate for the generalization *Nearly everything that has property $\alpha$ also has property $\beta$*. Note that $\pi^*(\neg\beta|\alpha)$ equals $\pi(\neg\beta|\alpha)$ unless $\pi(\alpha) = 0$, in which case $\pi^*(\neg\beta|\alpha)$ equals zero.

In practice, we seldom know the exact value of $\pi^*(\neg\beta|\alpha)$. Instead, what we know is some estimate $\pi^*(\neg\beta|\alpha)_{\text{est}}$ of the exception rate. Typically,

$$\pi^*(\neg\beta|\alpha) = \psi\pi^*(\neg\beta|\alpha)_{\text{est}}, \quad (3)$$

where $\psi$ is some unknown positive number which we hope is not very far from one. The unknown value of $\psi$ is not fixed, but may vary from one generalization to another.

Let $\delta$ be a small positive number. Let $\delta^1, \delta^2, \ldots$ and $\delta^\infty = 0$ be a collection of thresholds that have been adopted.

A *thresholded generalization* is a syntactic object having the form

$$\alpha \Rightarrow^k \beta$$

where $\alpha$ and $\beta$ are any properties in $\mathcal{L}$ and where $k$ is any positive integer or $\infty$. The meaning of $\alpha \Rightarrow^k \beta$ is that

$$\pi^*(\neg\beta|\alpha)_{\text{est}} \leq \delta^k. \quad (4)$$

Thus, from Eqs. 3 and 4, it is seen that $\alpha \Rightarrow^k \beta$ means that

$$1 - \pi(\beta|\alpha) \leq \psi\delta^k \quad (5)$$

where $\psi$ is an unknown positive number which we hope is not very far from one.

### 2.1  ALTERNATIVE INTERPRETATION

As discussed above, propositions in $\mathcal{L}$ are interpreted here as properties of objects and $\pi(\beta|\alpha)$ is interpreted as the conditional probability of an object having the property $\beta$ given that it has the property $\alpha$. Alternatively, propositions in $\mathcal{L}$ may be interpreted as assertions of fact and $\pi(\beta|\alpha)$ may be interpreted as the conditional probability of the assertion $\beta$ being true given that the assertion $\alpha$ is true. The logic presented in this paper is appropriate for either of the above interpretations.

Thresholded generalizations resemble conditional statements in Adams' (1966, 1975, 1986) logic of conditionals. An Adams conditional is a syntactic object of the form $\alpha \to \beta$ where $\alpha$ and $\beta$ are are propositions. Its meaning is that $\pi(\beta|\alpha)$ is close to one, where $\alpha$ and $\beta$ are interpreted as assertions.

---

[2] The convention expressed by Eq. 2 is controversial. Thus, Adams adopted this convention in the papers (Adams, 1966, 1986) but not in his book (Adams, 1975). McGee (1994) has proposed an alternative to this convention. The convention expressed by Eq. 2 was adopted in this paper for the following reason: The logic presented in this paper is intended for use in applications where $\pi(\alpha) = 0$ only if the set of objects having property $\alpha$ is empty. Consequently, Eq. 2 guarantees that $\pi(\beta|\alpha) = 1$ if and only if the objects having property $\alpha$ are a subset of the objects having property $\beta$.

## 2.2 GOALS

This paper has two major goals. The first goal is to formulate a nonmonotonic entailment relation, to be called *entailment in probability*, for thresholded generalizations. As a stepping stone toward the definition of entailment in probability, a monotonic entailment relation called *entailment with certainty* will be defined.

The second goal is to find a procedure for ascertaining whether any given collection of premises and any given conclusion satisfy the definition of entailment in probability.

## 3 ENTAILMENT CRITERIA

Classical logic uses the following criterion for entailment.

**Criterion 1 (Classical)** *For a set of premises to entail a conclusion, it should be impossible for the conclusion to be false when the premises are true. Thus, every model of the premises should be a model of the conclusion. In short, any entailed conclusion should be* trustworthy.

Any entailment relation that satisfies this classical criterion will necessarily be monotonic.

Since a nonmonotonic entailment relation cannot satisfy the classical criterion for entailment, what criterion should it satisfy instead? Instead of adopting some new criterion that is unrelated to the classical criterion, it makes sense to adopt a modified form of the classical criterion. The following weakened form of the classical criterion is proposed as a sensible criterion for nonmonotonic entailment.

**Criterion 2 (Modified Classical)** *For a set of premises to entail a conclusion, it should be* improbable *that the conclusion would be false when the premises are true. Thus,* nearly *every model of the premises should be a model of the conclusion. In short, any entailed conclusion should be* probabilistically trustworthy.

Most of our everyday reasoning does not satisfy the classical criterion for entailment. We continually leap to conclusions that are not fully certain given the premises from which we are reasoning. Indeed our reasoning would be virtually paralyzed if we used only the classical criterion for entailment. Perhaps, in our everyday reasoning, we frequently use the modified classical criterion.

As formulated above, the modified classical criterion is imprecise. It shall be one of the goals of this paper to rigorously and precisely define a nonmonotonic entailment relation that embodies the spirit of the modified classical criterion.

## 4 MONOTONIC ENTAILMENT

We will now work toward defining, first, a monotonic entailment relation and, later, a nonmonotonic entailment relation for thresholded generalizations. Recall that the thresholded generalization $\alpha \Rightarrow^k \beta$ means that

$$1 - \pi(\beta|\alpha) \leq \psi\delta^k$$

where $\delta$ is a positive number close to zero and $\psi$ is a positive number that we hope is not very far from one. The definitions of these entailment relations will be idealizations in two ways. First, because $\delta$ is close to zero, the definitions will involve looking at asymptotes as $\delta$ goes to zero. Second, because the value of $\psi$ is unknown, the definitions will not depend upon $\psi$ having any particular value or range of values.

### 4.1 MODELS

**Definition 3** *A* model *is a probability function on $\mathcal{L}$. Let $\mathcal{M}$ denote the set of all models. Suppose that $\pi$ is a model and that $\psi > 0$ and $\delta > 0$. Then $\pi$ satisfies the thresholded generalization $\alpha \Rightarrow^k \beta$ under the parameters $(\psi, \delta)$ if and only if*

$$1 - \pi(\beta|\alpha) \leq \psi\delta^k.$$

*Define $\mathcal{M}_{\psi,\delta}(\alpha \Rightarrow^k \beta)$ to be the set of models that satisfy $\alpha \Rightarrow^k \beta$ under the parameters $(\psi, \delta)$.*

**Notation 4** *Throughout the remainder of this paper, let $\mathcal{A}$ denote the set of $m \geq 0$ thresholded generalizations:*

$$\mathcal{A} = \{\alpha_1 \Rightarrow^{k_1} \beta_1, \ldots, \alpha_m \Rightarrow^{k_m} \beta_m\}.$$

**Definition 5** *Let $\vec{\Psi}$ denote the vector $(\psi_1, \ldots, \psi_m)$. Call $\vec{\Psi}$ a* positive parameter vector *if $\psi_i > 0$ for $i = 1, \ldots, m$. For any positive parameter vector $\vec{\Psi}$ and any $\delta > 0$, define $\mathcal{M}_{\vec{\Psi},\delta}(\mathcal{A})$ to be the intersection of $\mathcal{M}$ with each of $\mathcal{M}_{\psi_i,\delta}(\alpha_i \Rightarrow^{k_i} \beta_i)$, $i = 1, \ldots, m$.*

### 4.2 CONSISTENCY

**Abbreviation 6** *Let $S(\delta)$ be any statement involving $\delta$. Throughout this paper, (a) and (b) will be used as abbreviations for (c).*

(a) *For small $\delta$, $S(\delta)$.*

(b) *$S(\delta)$ for small $\delta$.*

(c) *There exists a $\Delta > 0$ such that, for all positive $\delta \leq \Delta$, $S(\delta)$.*

**Proposition/Definition 7** *Either* (a) *or* (b) *is true.*

(a) *For every positive parameter vector $\vec{\Psi}$ and every positive $\delta$, $\mathcal{M}_{\vec{\Psi},\delta}(\mathcal{A})$ is not empty. In this case, $\mathcal{A}$ is said to be* consistent.

60    Bamber**(b)** *For every positive parameter vector $\vec{\Psi}$, $\mathcal{M}_{\vec{\Psi},\delta}(\mathcal{A})$ is empty for small $\delta$. In this case, $\mathcal{A}$ is said to be inconsistent.*

(Space limitations preclude giving proofs here. Proofs will appear in the full version of this paper.)

### 4.3 ENTAILMENT WITH CERTAINTY

**Definition 8 (Standard Definition)** *Suppose $g(\cdot)$ and $h(\cdot)$ are non-negative functions defined on an open interval $(0,a)$. Suppose that there exists a $\psi > 0$ such that, for small $\delta$,*

$$h(\delta) \leq \psi g(\delta).$$

*Then,*

$$h(\delta) = O[g(\delta)].$$

**Definition 9** *$\mathcal{A}$ entails $\gamma \Rightarrow^j \zeta$ with certainty (denoted $\mathcal{A} \models_c \gamma \Rightarrow^j \zeta$) if either* **(a)** *or* **(b)**.

**(a)** *$\mathcal{A}$ is consistent and, for every positive parameter vector $\vec{\Psi}$,*

$$\sup_{\pi \in \mathcal{M}_{\vec{\Psi},\delta}(\mathcal{A})} [1 - \pi(\zeta|\gamma)] = O(\delta^j). \qquad (6)$$

**(b)** *$\mathcal{A}$ is inconsistent.*

**Proposition 10** *$\mathcal{A}$ entails $\gamma \Rightarrow^j \zeta$ with certainty if, for every positive parameter vector $\vec{\Psi}$, there exists a $\psi > 0$ such that*

$$\mathcal{M}_{\vec{\Psi},\delta}(\mathcal{A}) \subseteq \mathcal{M}_{\psi,\delta}(\gamma \Rightarrow^j \zeta) \qquad (7)$$

*for small $\delta$.*

Loosely speaking, Proposition 10 says that $\mathcal{A}$ entails $\gamma \Rightarrow^j \zeta$ with certainty if and only if *every* model that satisfies $\mathcal{A}$ also satisfies $\gamma \Rightarrow^j \zeta$. Also, this proposition implies that entailment with certainty is monotonic.

### 4.4 ADAMS' LOGIC OF CONDITIONALS

Entailment with certainty is closely related to entailment in Adams' (1966) logic of conditionals. Using results of Adams including Meta-metatheorems 1 and 2 of Adams (1986), the following result can be proven.

**Theorem 11** *Suppose that $k_1, \ldots, k_m$, where $m \geq 0$, are all finite. If*

$$j \leq \min\{\infty, k_1, \ldots, k_m\},$$

*then $\mathcal{A}$ entails $\gamma \Rightarrow^j \zeta$ with certainty if and only if, in Adams' (1966) logic of conditionals,*

$$\{\alpha_1 \to \beta_1, \ldots, \alpha_m \to \beta_m\}$$

*entails $\gamma \to \zeta$.*

Theorem 11 does not provide a complete characterization of entailment with certainty. For example,

$$\{t \Rightarrow^1 \alpha,\ \neg\alpha \Rightarrow^1 \beta\} \models_c t \Rightarrow^2 \alpha \vee \beta,$$

but that fact cannot be derived from Theorem 11.

Entailment with certainty will not be fully characterized here. It has been introduced primarily as a stepping stone toward the definition of a nonmonotonic entailment relation to be called *entailment in probability*.

## 5 NONMONOTONIC ENTAILMENT

A nonmonotonic entailment relation for thresholded generalizations will now be constructed by taking the requirements for monotonic entailment and making them less stringent. Loosely speaking, instead of requiring a conclusion to be satisfied by *every* model that satisfies the premises, it will only be required that the conclusion be satisfied by *nearly every* model that satisfies the premises. This strategy for defining a nonmonotonic entailment relation is inspired by the random-worlds and random-structures methods (Bacchus, Grove, Halpern & Koller, 1992, 1993, 1996; Grove, Halpern & Koller, 1994, 1996a, 1996b).

### 5.1 PROBABILITY ORDER

The concept of *probability order* and the notation $O_p(\cdot)$ were introduced by Mann and Wald (1943) and elaborated by Chernoff (1956) and Pratt (1959). A nice tutorial is presented by Bishop, Fienberg and Holland (1975, Sections 14.2 & 14.4).

**Definition 12** *Suppose that, for each $\delta$ in an open interval $(0,a)$, $X(\delta)$ is a non-negative random variable and $g(\delta)$ is non-negative number. Suppose that, for each $\eta > 0$, there exists a $\psi > 0$ such that, for small $\delta$,*

$$\Pr[X(\delta) \leq \psi g(\delta)] \geq 1 - \eta.$$

*Then,*

$$X(\delta) = O_p[g(\delta)].$$

**Proposition 13** *Define the inverse cumulative distribution function of the non-negative random variable $X(\delta)$ as follows. For each $\eta > 0$, let*

$$cdf^{-1}_{X(\delta)}(1-\eta) = \inf\{x \geq 0 : \Pr[X(\delta) \leq x] \geq 1 - \eta\}$$

*Then, $X(\delta) = O_p[g(\delta)]$ if and only if, for each $\eta > 0$,*

$$cdf^{-1}_{X(\delta)}(1-\eta) = O[g(\delta)].$$

### 5.2 MODELS AS VECTORS

Following Nilsson (1986) and Paris (1994, pp. 13–14), probability functions may be represented as points in a finite-dimensional space. Recall that the language $\mathcal{L}$ contains $2^r$ atoms. Therefore, to specify a probability function (*i.e.*, model) $\pi$ on $\mathcal{L}$, it suffices to specify the value of $\pi(at)$ for each atom $at \in \mathcal{L}$. Therefore, any model may be represented by a vector in $2^r$-dimensional space. The components of such a vector will all be non-negative and will sum to one. The set of models $\mathcal{M}$ is represented by a $(2^r - 1)$-dimensional polytope, namely, the convex hull of the $2^r$ unit axis vectors.

These two views of models will be used interchangeably throughout this paper. Sometimes a model will be considered to be a probability function and sometimes a vector.

If each model in $\mathcal{M}_{\vec{\Psi},\delta}(\mathcal{A})$ is considered to be a vector, then it can be shown that the set of vectors $\mathcal{M}_{\vec{\Psi},\delta}(\mathcal{A})$ is a convex polytope in $2^r$-dimensional space. Its dimension is at most $2^r - 1$.

**Definition 14** *If the polytope $\mathcal{M}_{\vec{\Psi},\delta}(\mathcal{A})$ is not empty, let $\mu^{\mathcal{A}}_{\vec{\Psi},\delta}$ denote the uniform probability measure on $\mathcal{M}_{\vec{\Psi},\delta}(\mathcal{A})$. Let $\Pi^{\mathcal{A}}_{\vec{\Psi},\delta}$ be a random vector that has distribution $\mu^{\mathcal{A}}_{\vec{\Psi},\delta}$.*

Because vectors are used here to represent models, $\Pi^{\mathcal{A}}_{\vec{\Psi},\delta}$ may be viewed as being a random model. Thus, we are dealing with two levels of probabilities. At the lower level, each model is a probability function. At the upper level, there is a probability distribution over models.[3]

## 5.3 ENTAILMENT IN PROBABILITY

We want to define entailment in probability so that, loosely speaking, $\mathcal{A}$ entails $\gamma \Rightarrow^j \zeta$ in probability if and only if *nearly every* model that satisfies each thresholded generalization in $\mathcal{A}$ also satisfies $\gamma \Rightarrow^j \zeta$.

When is one justified in asserting that *nearly every* model in $\mathcal{M}_{\vec{\Psi},\delta}(\mathcal{A})$ has some particular attribute $\xi$? There are two cases to consider. First, suppose that $\mathcal{A}$ is consistent and, therefore, $\mathcal{M}_{\vec{\Psi},\delta}(\mathcal{A})$ is not empty. If the probability that the random model $\Pi^{\mathcal{A}}_{\vec{\Psi},\delta}$ has attribute $\xi$ is close to one, then it is reasonable to say that nearly every model in $\mathcal{M}_{\vec{\Psi},\delta}(\mathcal{A})$ has attribute $\xi$.

Second, suppose that $\mathcal{A}$ inconsistent and, therefore, $\mathcal{M}_{\vec{\Psi},\delta}(\mathcal{A})$ is empty for small $\delta$. Then, *every* model in $\mathcal{M}_{\vec{\Psi},\delta}(\mathcal{A})$ has the attribute $\xi$.

In the following definition, since $\Pi^{\mathcal{A}}_{\vec{\Psi},\delta}$ is a random model, the conditional probability $\Pi^{\mathcal{A}}_{\vec{\Psi},\delta}(\zeta|\gamma)$ is a random variable.

**Definition 15** $\mathcal{A}$ *entails* $\gamma \Rightarrow^j \zeta$ *in probability (denoted $\mathcal{A} \models_p \gamma \Rightarrow^j \zeta$) if either* **(a)** *or* **(b)**.

**(a)** $\mathcal{A}$ *is consistent and, for every positive parameter vector $\vec{\Psi}$,*

$$1 - \Pi^{\mathcal{A}}_{\vec{\Psi},\delta}(\zeta|\gamma) = O_p(\delta^j). \qquad (8)$$

**(b)** $\mathcal{A}$ *is inconsistent.*

**Proposition 16** $\mathcal{A}$ *entails* $\gamma \Rightarrow^j \zeta$ *in probability if and only if either* **(a)** *or* **(b)**.

**(a)** $\mathcal{A}$ *is consistent and, for every positive parameter vector $\vec{\Psi}$ and every $\eta > 0$, there exists a $\psi > 0$ such that*

$$\mu^{\mathcal{A}}_{\vec{\Psi},\delta}[\mathcal{M}_{\psi,\delta}(\gamma \Rightarrow^j \zeta)] \geq 1 - \eta \qquad (9)$$

*for small $\delta$.*

**(b)** $\mathcal{A}$ *is inconsistent.*

Loosely speaking, Proposition 16 says that $\mathcal{A}$ entails $\gamma \Rightarrow^j \zeta$ in probability if and only if *nearly every* model that satisfies $\mathcal{A}$ also satisfies $\gamma \Rightarrow^j \zeta$. Not surprisingly, it turns out that entailment in probability is nonmonotonic.

Definition 15 and Proposition 16 which define and characterize entailment in probability are stochastic analogs of Definition 9 and Proposition 10 which define and characterize entailment with certainty. Thus, Eq. 8 is the stochastic analog of Eq. 6 and Eq. 9 is the stochastic analog of Eq. 7.

## 5.4 ROBUSTNESS

In essence, the definition of entailment in probability is Bayesian.

Recall that the set $\mathcal{M}$ of all models is a polytope in $2^r$-dimensional space. The dimension of $\mathcal{M}_{\vec{\Psi},\delta}(\mathcal{A})$ may decrease as $\delta$ goes to zero. Let $\mathcal{F}$ denote the lowest-dimension face of $\mathcal{M}$—see Webster (1994, pp. 79–80) for the definition of *face*—that eventually contains $\mathcal{M}_{\vec{\Psi},\delta}(\mathcal{A})$ as $\delta$ goes to zero. Let $\mu$ denote the uniform distribution on $\mathcal{F}$. Then, for small $\delta$, Eq. 9 may be rewritten with a posterior probability on the left-hand side:

$$\mu[\mathcal{M}_{\psi,\delta}(\gamma \Rightarrow^j \zeta)|\mathcal{M}_{\vec{\Psi},\delta}(\mathcal{A})] \geq 1 - \eta. \qquad (10)$$

---

[3]This method of adopting a uniform distribution over models is similar to the random-structures method (Bacchus, Grove, Halpern & Koller, 1992; Grove, Halpern & Koller, 1996a, 1996b). In this method, a *structure* is an allocation of $N \geq 1$ *indistinguishable* objects among $2^r$ atoms. Thus, a structure may be represented by a vector of $2^r$ non-negative integers that sum to $N$. Each structure is assigned equal probability. Let $X_N$ denote a randomly selected structure. As $N \to \infty$, the distribution of $X_N/N$ approaches a uniform distribution. The goal of the random-structures method is to find the asymptotic conditional probability of a sentence $\varphi$ being satisfied given that a sentence $\theta$ was satisfied.

If the definition of entailment in probability were based upon Eq. 10, how robust would the definition be over changes in the prior distribution $\mu$? If $\mu$ were not uniform, would the entailment relation change?

It can be shown that, if $\mu$ is quasi-uniform in the sense that its density has a lower bound greater than zero and an upper bound less than infinity, then the entailment relation will be unchanged.

In at least some cases, choices of $\mu$ that are not quasi-uniform leave the entailment relation unchanged. An example can be constructed as follows. Imagine extending the propositional language $\mathcal{L}$ from having $r$ primitive propositions to a language $\mathcal{L}^*$ having a larger number $r+s$ of primitive propositions. Let $\mathcal{M}^*$ denote the set of all models for $\mathcal{L}^*$. Any model in $\mathcal{M}^*$ may be converted to a model in $\mathcal{M}$ by the appropriate projection from $2^{r+s}$-dimensional space to $2^r$-dimensional space. This projection carries a uniform distribution on $\mathcal{M}^*$ to a Dirichlet distribution on $\mathcal{M}$ and this latter distribution is not quasi-uniform. But, it follows from Theorem 27 below that entailment in probability is language independent. Therefore, if this particular Dirichlet distribution were chosen as a prior distribution on $\mathcal{M}$, the resulting entailment relation would be unchanged.

### 5.5 DEGREE OF RARITY

The next definition defines *degree of rarity* in terms of entailment with certainty and the succeeding two theorems relate degree of rarity to consistency and to entailment in probability.

**Definition 17** *For any property $\rho \in \mathcal{L}$, define its degree of rarity under $\mathcal{A}$ by:*

$$r^\circ_\mathcal{A}(\rho) = \max\left[\{0\} \cup \{k \leq \infty : \mathcal{A} \models_c t \Rightarrow^k \neg\rho\}\right].$$

**Theorem 18** *If $\mathcal{A}$ is consistent, then $r^\circ_\mathcal{A}(t) = 0$. On the other hand, if $\mathcal{A}$ is inconsistent, then $r^\circ_\mathcal{A}(t) = \infty$.*

**Theorem 19** $\mathcal{A} \models_p \gamma \Rightarrow^j \zeta$ *if and only if*

$$r^\circ_\mathcal{A}(\gamma \wedge \neg\zeta) \geq r^\circ_\mathcal{A}(\gamma) + j.^4 \qquad (11)$$

Theorem 19 is the key result upon which later theorems concerning entailment in probability are based. Theorem 19 is proved by showing that Eq. 8 of Definition 15 is equivalent to a statement about the support function (Webster, 1994) of the polytope $\mathcal{M}_{\bar{\Psi},\delta}(\mathcal{A})$.

## 6 DECISION PROCEDURE

### 6.1 DEPTH ENTAILMENT DEFINED

Depth entailment appears, at first, to be a new kind of entailment but it turns out to be equivalent to entailment in probability.

---

[4]This paper uses the convention that $\infty \geq h + j$ if either or both of $h$ and $j$ is $\infty$.

**Definition 20** *A function $d$ that maps $\mathcal{L}$ into the non-negative integers plus $\infty$ is called a depth function if, for every $\alpha, \beta \in \mathcal{L}$,*

- $d(\alpha) \geq d(\beta)$ *if* $\alpha \models \beta$.
- $d(\alpha \vee \beta) = \min\{d(\alpha), d(\beta)\}$.
- $d(f) = \infty$.

*A depth function $d$ satisfies a thresholded generalization $\gamma \Rightarrow^j \zeta$ if and only if*

$$d(\gamma \wedge \neg\zeta) \geq d(\gamma) + j.^5 \qquad (12)$$

*A depth function satisfies a collection of thresholded generalizations if and only if it satisfies every member of the collection.*

**Proposition 21** *There exists a unique depth function, denoted $d_\mathcal{A}$, that satisfies $\mathcal{A}$ and such that, if $d$ is any other depth function that satisfies $\mathcal{A}$, then $d_\mathcal{A}(\gamma) \leq d(\gamma)$ for every $\gamma \in \mathcal{L}$. The quantity $d_\mathcal{A}(\gamma)$ is called the* exception depth *of $\gamma$ under $\mathcal{A}$.*

**Definition 22** $\mathcal{A}$ *depth entails* $\gamma \Rightarrow^j \zeta$ *(denoted $\mathcal{A} \models_d \gamma \Rightarrow^j \zeta$) if and only if $d_\mathcal{A}$ satisfies $\gamma \Rightarrow^j \zeta$.*

### 6.2 TESTING FOR DEPTH ENTAILMENT

**Definition 23** *The* exceptions to $\mathcal{A}$ at depth $d$ and beyond, *denoted $\xi_\mathcal{A}(d)$, will be defined inductively. For $d = 0, -1, -2, \cdots, -\infty$, let $\xi_\mathcal{A}(d) = t$. Next, suppose that $d$ is a positive integer and that $\xi_\mathcal{A}(d')$ has been defined for every integer $d' < d$. Then, for $i = 1, \ldots, m$, let*

$$\nu_i(d) = \begin{cases} \alpha_i \wedge \neg\beta_i & \text{if } \alpha_i \models \xi_\mathcal{A}(d - k_i); \\ f & \text{otherwise}. \end{cases}$$

*Then, let*

$$\xi_\mathcal{A}(d) = f \vee \nu_1(d) \vee \cdots \vee \nu_m(d).$$

**Proposition/Definition 24** $\xi_\mathcal{A}(d') \models \xi_\mathcal{A}(d)$ *whenever $d' > d$. There exists a smallest non-negative integer $D$ such that $\xi_\mathcal{A}(D') \models = \models \xi_\mathcal{A}(D)$ whenever $D' > D$. Moreover, if*

$$k = \max\left[\{1\} \cup \{K \neq \infty : K = k_i \text{ for some } i\}\right],$$

*then $D$ is the smallest non-negative integer such that*

$$\xi_\mathcal{A}(D) \models = \models \xi_\mathcal{A}(D + k). \qquad (13)$$

*Define*

$$\xi_\mathcal{A}(\infty) = \xi_\mathcal{A}(D).$$

**Proposition 25** *For any property $\rho \in \mathcal{L}$,*

$$d_\mathcal{A}(\rho) = \max\{d \leq \infty : \rho \models \xi_\mathcal{A}(d)\}.$$

---

[5]Note the similarity to Eq. 11.

Consequently, it is straightforward to test whether $\mathcal{A}$ depth entails $\gamma \Rightarrow^j \zeta$. Follow the procedure in Definition 23 to find $\xi_\mathcal{A}(0), \xi_\mathcal{A}(1), \ldots$. Stop the procedure when an integer $D$ that satisfies Eq. 13 is found. Then, use Proposition 25 to find $d_\mathcal{A}(\gamma \wedge \neg\zeta)$ and $d_\mathcal{A}(\gamma)$. Finally, check whether Eq. 12 holds.

## 6.3  TESTING FOR ENTAILMENT IN PROBABILITY

Exception depth and degree of rarity are the same thing.

**Proposition 26** *For any property $\rho \in \mathcal{L}$,*

$$d_\mathcal{A}(\rho) = r^\circ_\mathcal{A}(\rho).$$

Therefore, entailment in probability and depth entailment are the same thing.

**Theorem 27** $\mathcal{A} \models_p \gamma \Rightarrow^j \zeta$ *if and only if* $\mathcal{A} \models_d \gamma \Rightarrow^j \zeta$.

Thus, the procedure for ascertaining depth entailment is also a procedure for ascertaining entailment in probability.

## 7  RELATION TO SYSTEM-$Z^+$

Goldszmidt and Pearl (1991, 1992; Pearl, 1994) formulated a System-$Z^+$ for reasoning with default rules having varying strengths. In System-$Z^+$, a rule has the form $\alpha \overset{k}{\to} \beta$ where $\alpha$ and $\beta$ are propositions and $k$ is a non-negative integer. According to Pearl (1994, p. 58), the meaning of $\alpha \overset{k}{\to} \beta$ is that

$$\Pr(\beta|\alpha) \geq 1 - O(\epsilon^{k+1})$$

where $\epsilon$ is a positive number close to zero. Comparing the above equation with Eq. 5, it is evident that the default rule $\alpha \overset{k}{\to} \beta$ and the thresholded generalization $\alpha \Rightarrow^{k+1} \beta$ have essentially the same meaning.

Given any context $\Delta$ of default rules and an integer $k \geq 0$, Goldszmidt and Pearl (1992, Footnote 4) defined a consequence relation $\mathrel{\vdash\mkern-10mu\sim}^k_+$ between propositions. Thus, given the default rules $\Delta$, $\alpha \mathrel{\vdash\mkern-10mu\sim}^k_+ \beta$ means that $\beta$ follows from $\alpha$ and is endorsed with strength $k$.

Goldszmidt and Pearl constructed the consequence relations $\mathrel{\vdash\mkern-10mu\sim}^k_+$, $k = 0, 1, \ldots$, by applying an intuitively appealing principle that might be called *minimization of surprise*. Default rules may be interpreted as saying that certain things are surprising. Thus, the default rule $\alpha \overset{k}{\to} \beta$ implies that, if the proposition $\alpha$ is observed, then it is surprising to also observe the proposition $\neg\beta$ and, the larger $k$ is, the greater the surprise. Roughly speaking, Goldszmidt and Pearl showed that there is a way of minimizing the surprise attached to every proposition while still maintaining the surprise required by the default rules. They then defined $\gamma \mathrel{\vdash\mkern-10mu\sim}^j_+ \zeta$ to mean that, having observed the proposition $\gamma$ in this minimally surprising condition, one would be surprised to degree $j$ to also observe the proposition $\neg\zeta$.

System-$Z^+$ is closely related to depth entailment. Recall that

$$\mathcal{A} = \{\alpha_1 \Rightarrow^{k_1} \beta_1, \ldots, \alpha_m \Rightarrow^{k_m} \beta_m\}.$$

Let

$$\mathcal{A}_{Z^+} = \{\alpha_1 \overset{k_1-1}{\to} \beta_1, \ldots, \alpha_m \overset{k_m-1}{\to} \beta_m\}.$$

**Theorem 28** *Assume that $k_1, \ldots, k_m$ and $j$ are all positive and finite. Assume further that $\alpha_1, \ldots, \alpha_m$ and $\beta_1, \ldots, \beta_m$ and $\gamma$ and $\zeta$ are all not equivalent to $f$. Finally, assume that, for any proposition $\rho$, $\mathcal{A} \models_d t \Rightarrow^\infty \neg\rho$ only if $\rho \models f$.[6] Then, the following two statements are equivalent.*

(a) $\mathcal{A} \models_d \gamma \Rightarrow^j \zeta$.

(b) *Given the default rules $\mathcal{A}_{Z^+}$ as context, $\gamma \mathrel{\vdash\mkern-10mu\sim}^{j-1}_+ \zeta$.*

The basis for the above theorem is that Goldszmidt and Pearl's (1991) Theorem 3 and Corollary 2 imply that their ranking function $\kappa^+$ is essentially equivalent to this paper's depth function $d_\mathcal{A}$. Thus, this paper's definition of depth entailment is essentially an application of the principle of minimization of surprise.

In view of the close relation between entailment in probability, depth entailment, and System-$Z^+$, readers wishing to know about the properties of inference in these systems, the complexity of inference, and to see examples should consult Goldszmidt and Pearl(1991, 1992).

### 7.1  WHAT'S NEW IN THIS PAPER?

This paper started by proposing a symbolic language having a syntax and semantics similar to that of System-$Z^+$. The paper went on to develop the depth entailment relation which, except for special cases, is equivalent to the consequence relations in System-$Z^+$. So, what is new in this paper?

There are two new developments. The first new development was that entailment in probability was defined (Definition 15) so that it embodied the modified classical criterion for entailment. This criterion asserts a clearly desirable feature for a nonmonotonic entailment relation: Its conclusions should be probabilistically trustworthy meaning that, given true premises, any entailed conclusions should be unlikely to be false. Definition 15 guarantees that conclusions entailed in probability are probabilistically trustworthy.

---

[6]The purpose of this requirement is to guarantee that $\mathcal{A}_{Z^+}$ is consistent as defined by Goldszmidt and Pearl (1991, Definition 4).

But, it is not enough to merely define a nonmonotonic entailment relation whose conclusions are probabilistically trustworthy. Also needed is a decision procedure for ascertaining whether any given conclusion is entailed by any given collection of premises. However, from the definition of entailment in probability, it's not obvious what a decision procedure for it would be.

In marked contrast to entailment in probability, the definition of depth entailment leads straightforwardly to a decision procedure. However, the drawback to depth entailment was that it could only be justified by the minimization-of-surprise principle. Although this principle is intuitively appealing, the principle itself is hard to justify. Moreover, this principle does not seem to guarantee the probabilistic trustworthiness of entailed conclusions.

The second new development in this paper was the demonstration (Theorem 27) that entailment in probability is identical to depth entailment. By asserting the identity of these two types of entailment, Theorem 27 simultaneously provides the missing decision procedure for entailment in probability and a stronger justification for depth entailment.

In addition, given the close relation between System-$Z^+$ and depth entailment, Theorem 27 also puts System-$Z^+$ on a more sound footing and shows that its conclusions are probabilistically trustworthy.

## Acknowledgments

In thinking about the issues discussed in this paper, I have been greatly helped by the comments of Ernest Adams and Joseph Halpern.